\documentclass{vgtc}
\usepackage[utf8]{inputenc}
\usepackage{amssymb}
\usepackage{algorithm}
\usepackage[noend]{algpseudocode}
\usepackage{graphicx}  
\usepackage{amsmath}
\usepackage{color}
\usepackage{amsmath}

\ifpdf
  \pdfoutput=1\relax                   
  \usepackage{graphicx}                
  \DeclareGraphicsExtensions{.pdf,.png,.jpg,.jpeg} 
\else
  \usepackage{graphicx}                
  \DeclareGraphicsExtensions{.eps}     
\fi%

\graphicspath{{figures/}{pictures/}{images/}{./}} 

\usepackage{microtype}                 
\PassOptionsToPackage{warn}{textcomp}  
\usepackage{textcomp}                  
\usepackage{mathptmx}                  
\usepackage{times}                     
\renewcommand*\ttdefault{txtt}         
\usepackage{cite}                      
\usepackage{tabu}                      
\usepackage{booktabs}                  

\usepackage{enumitem}
\setlist[itemize]{leftmargin=*, topsep=0pt, itemsep=0pt, parsep=0pt, partopsep=0pt}
\setlist[enumerate]{leftmargin=*, topsep=0pt, itemsep=0pt, parsep=0pt, partopsep=0pt}

\onlineid{0}

\vgtccategory{Research}

\vgtcinsertpkg

\title{\textit{EmbeddingTree}: Hierarchical Exploration of Entity Features in Embedding}

\author{Yan Zheng\thanks{e-mail: yazheng@visa.com}
\and Junpeng Wang\thanks{e-mail:junpenwa@visa.com} 
\and Chin-Chia Michael Yeh\thanks{e-mail:miyeh@visa.com}
\and Yujie Fan\thanks{e-mail:yufan@visa.com}
\and Huiyuan Chen\thanks{e-mail:hchen@visa.com}
\and Liang Wang\thanks{e-mail:liawang@visa.com}
\and Wei Zhang\thanks{e-mail:wzhan@visa.com}}
\affiliation{\scriptsize Visa Research}

\teaser{
  \centering
  \includegraphics[width=\textwidth]{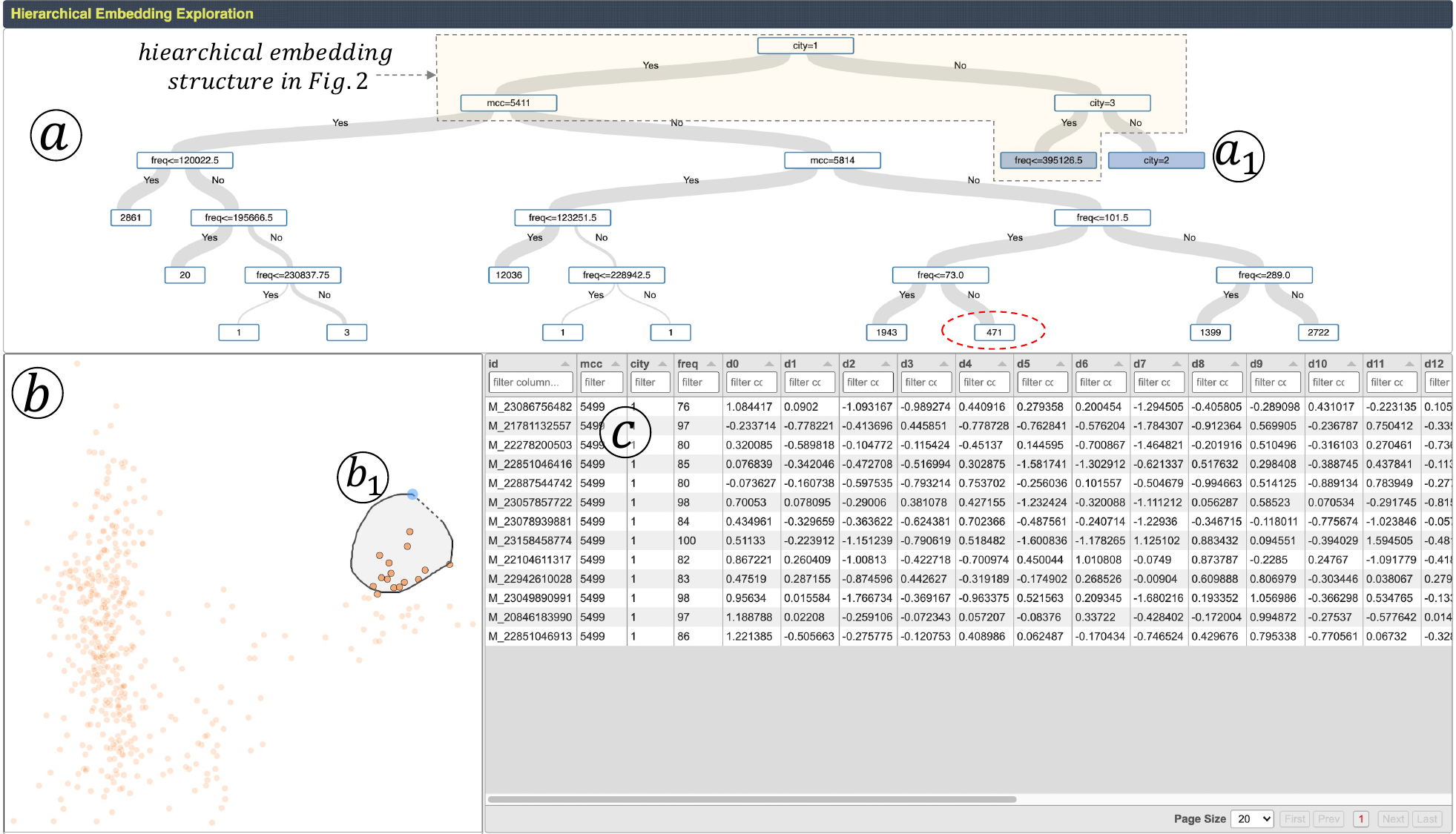}
  \vspace{-0.25in}
  \caption{Our visualization system with three coordinated views. The Tree view (a) presents the structure of an embedding tree and data distributions across branches. The Dimensionality Reduction view (b) shows data entities from a selected tree node and reveals the cluster information. The Data Table (c) shows details of the selected data entities and enables users to search/filter/sort them.}
  \label{fig:vis}
}

\abstract{
Embedding learning transforms discrete data entities into continuous numerical representations, encoding features/properties of the entities.
Despite the outstanding performance reported from different embedding learning algorithms, few efforts were devoted to structurally interpreting how features are encoded in the learned embedding space. 
This work proposes \textit{EmbeddingTree}, a hierarchical embedding exploration algorithm that relates the semantics of entity features with the less-interpretable embedding vectors. 
An interactive visualization tool is also developed based on \textit{EmbeddingTree} to explore high-dimensional embeddings.
The tool helps users discover nuance features of data entities, perform feature denoising/injecting in embedding training, and generate embeddings for unseen entities. We demonstrate the efficacy of \textit{EmbeddingTree} and our visualization tool through embeddings generated for industry-scale merchant data and the public 30Music listening/playlists dataset.
}

\begin{document}

\maketitle

\section{Introduction}
Embedding learning has emerged as a powerful technique for analyzing text documents~\cite{mikolov2013efficient,camacho2018word} and other data types~\cite{tshitoyan2019unsupervised,grbovic2018real,wang2018billion,zhao2018learning,du2019}. It encapsulates relevant information into high-dimensional (HD) vectors that could be utilized for various downstream tasks. 
Often, there is no clear correspondence between raw features and individual embedding dimensions, and each dimension could entangle the semantics of multiple features.
This flexibility allows for embedding more features and better capturing the correlations between them.
However, it also eschews the interpretability of the embeddings.


Targeting on the interpretability problem of embeddings, multiple visual exploration solutions have been proposed~\cite{smilkov2016embedding, heimerl2020embcomp, liu2019latent, liu2017visual,berger2020visually}. These solutions explore embeddings by relating their coordinates' similarity to specific features or feature combinations. For example, a cluster of embeddings with similar spatial locations in \textit{embedding projector}~\cite{smilkov2016embedding} often corresponds to data entities with similar feature values. The correspondence helps to disclose the encoded semantics in the embedding. We consider this type of exploration \textit{parallel exploration}, as features are treated equally and explored in parallel. 

\begin{figure}[tbh]
\includegraphics[width=\columnwidth]{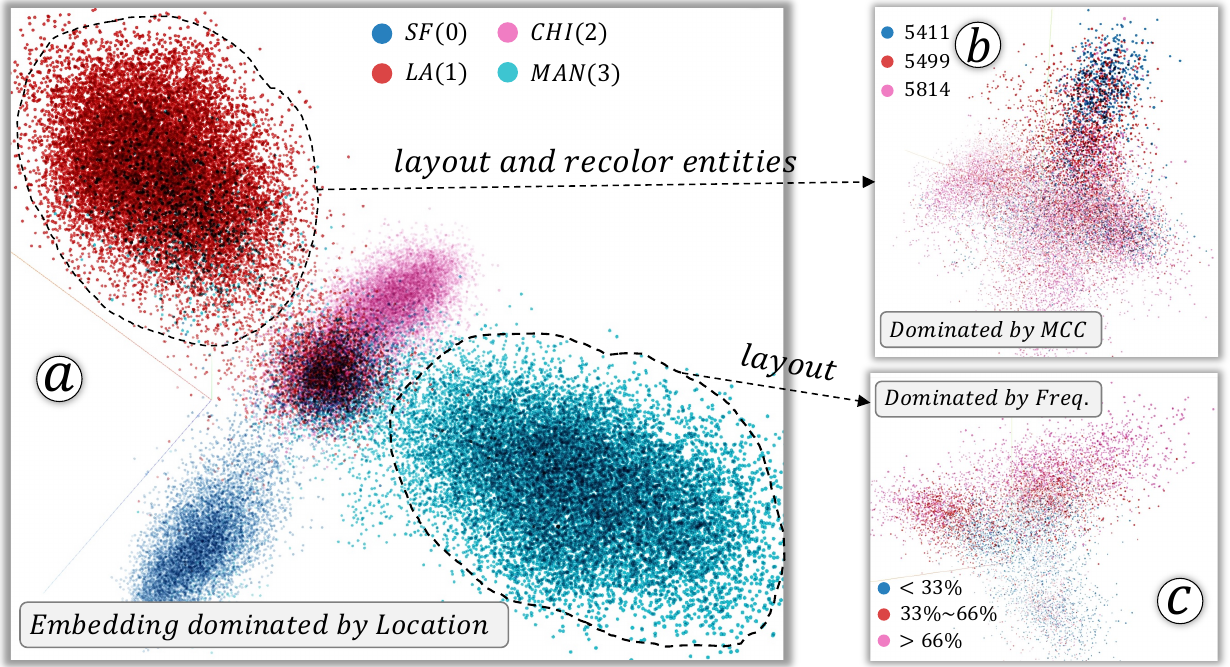}
  \caption{Hierarchical structure in the embedding. 
  (a) merchant embeddings are dominated by the location feature and separated into four clusters for the four cities. (b, c) For the merchants from Los Angeles (LA) and Manhattan (MAN), MCC and visiting frequency further dominate the embedding layouts, respectively. The tree branches highlighted in light-yellow in Figure~\ref{fig:vis}a show this hierarchy.}
  \label{fig:embedding}
\end{figure}

In some datasets, however, features are not of the same importance and should be explored in order or following a hierarchy. For example, in the merchant embeddings generated by Wang et al.~\cite{wang2021constrained} using credit-card transactions, merchants' \textit{location} is the feature that dominates the embedding the most, visiting \textit{frequency} is the second, and merchants' category code (\textit{MCC}) is the third. As shown in Figure~\ref{fig:embedding}a, when using PCA to project the merchant embeddings, the merchants get separated into four clear clusters, representing merchants from four different cities. Focusing on the red and cyan clusters (merchants from Los Angeles and Manhattan), \textit{MCC} and visiting \textit{frequency} dominate the layouts of the two local clusters respectively (Figure~\ref{fig:embedding}b, c). Existing parallel explorations cannot easily reveal this hierarchical feature dominance and could hinder the exploration process from discovering feature-level insights.


To accommodate the aforementioned problem, we introduce \textit{hierarchical exploration} in this work, where features form nested structures, and users can explore them in a top-down hierarchy. 
The hierarchy in Figure~\ref{fig:embedding} could be extracted as a tree structure, as highlighted in the light yellow region of Figure~\ref{fig:vis}a, guiding the exploration process.
The benefit of such exploration is twofold: (1) the hierarchy directly infers features' importance in the embedding; (2) the hierarchy provides explicit guidance and helps users drill down to embeddings with interested features more efficiently. 



In detail, we \textit{first} propose a Gaussian Mixture Model (GMM) based algorithm to generate an \textit{EmbeddingTree} from HD embeddings. The algorithm uses entity features to split the embedding space based on which feature can best separate the embedding in nested local regions through Bayesian Information Criterion (BIC). 
\textit{Next}, we develop a visualization tool (Figure~\ref{fig:vis}) to visualize the trained embeddings and organize them according to the hierarchical entity features in the \textit{EmbeddingTree}.
The tool employs interactive tree visualization, dimensionality reduction, and data table visualization to facilitate the hierarchical exploration of HD embeddings.
Our \textit{EmbeddingTree} algorithm and the visual exploration can help to analyze the hierarchical importance of entity features in the embeddings. As a result, users could discover the inconsistency between entity features and embeddings, as well as generate the embeddings for new entities with little structure information but rich entity features. 

To sum up, our contributions in this work are:
\begin{enumerate}
    \item We propose a GMM-based algorithm to extract a feature hierarchy from HD embeddings.
    \item We introduce a visual analytics tool that helps users effectively explore embedding data following the extracted hierarchy. 
\end{enumerate}
\section{Related Work}
\paragraph{Embedding Generation}
The embedding technology is popularized by the introduction of Word2Vec~\cite{MSC13} and GloVe \cite{pennington2014glove}. Both were proposed for words and rely on the distributional hypothesis of Harris~\cite{harris1954}, i.e., words occurring in similar contexts tend to have similar meanings. Word embedding generated from those methods captures the semantic and syntactic properties of words. 
Another popular group of embeddings are graph embeddings, which aim to represent graphs or graph nodes as HD vectors/embeddings. These embeddings capture the essential information from both the structure-level and node-level of graphs and help to boost the downstream tasks' performance, e.g., node classification~\cite{li2019learning}, recommendation~\cite{fan2019graph} and link prediction~\cite{wei2017cross}. With the fast evolution of deep learning techniques, more and more embedding learning techniques~\cite{cai2018comprehensive,goyal2018graph,rahman2019fairwalk,wang2017knowledge, hamilton2017inductive} have been proposed, and they have been used to generate embeddings for different types of data.

Our \textit{EmbedingTree} algorithm and the visualization tool can be applied to the embedding dataset with categorical/numerical features, regardless of the algorithm to generate the embeddings. For example, the merchant embeddings in Section~\ref{sec:merchant} were trained from credit-card transaction sequences, using the the Word2vec~\cite{mikolov2013distributed} algorithm; the user embeddings in Section~\ref{sec:music} were trained from the 30Music dataset, using the embedding learning framework proposed in~\cite{yeh2020towards}.

\paragraph{Embedding Visualization}
Existing embedding visualization works can roughly be categorized into three groups, focusing on \textit{semantic interpretation}, \textit{analogy reasoning}, and \textit{comparison} of embeddings. For the first group, Latent Space Cartography ~\cite{liu2019latent} and LatentVis~\cite{liu2020latentvis} disclose how semantics are encoded and expressed along certain directions of the HD latent/embedding space.
For the second group, Liu et al.~\cite{liu2017visual} study the analogy relation between pairs of word embeddings and optimize embedding generation algorithms by better preserving the relation. For the third group of embedding comparisons, EmbeddingVis~\cite{li2018embeddingvis} compares graph embeddings generated from different methods by linking multiple juxtaposed scatterplots. embComp~\cite{heimerl2020embcomp} quantifies the similarity of two embeddings through the number of shared local neighbors.
Recently, Embedding Comparator\cite{boggust2022embedding} presents a global comparison of embedding spaces alongside fine-grained inspection of local neighborhoods.

Most of these embedding visualizations explore embedding data without explicitly leveraging the relationship between entity features. Our work targets at this missing piece and strives to explore embeddings by following a hierarchy of data features distilled from the embedding data. This helps to structurally explore the embedding space and better understand the encoded semantics.


\section{Algorithm}


\label{sec:algorithm}

Our \textit{EmbeddingTree} construction is based on the traditional decision tree algorithm but uses two sets of data, one is the embedding data and the other is entity features. Entity features are used as the search criteria to split the embedding by efficiently computing an approximate Bayesian Information Criterion (BIC) for a Gaussian Mixture Model (GMM).  



\subsection{Embedding Tree Algorithm}
The input of \textit{EmbeddingTree} is a combination of an embedding dataset $\mathbf{X} = \{\mathbf{x}_1,...,\mathbf{x}_N \}$ ($\mathbf{x}_i \in \mathbb{R}^p$, $1 \leq  i \leq N$) and the corresponding feature set. For numerical and multi-categorical feature, we pre-process the features by (1) binning numerical features into categorical ones; (2) converting non-binary categorical features into binary ones by asking yes-no questions regarding to each of their categorical values. Finaly we have the binary feature sets: $\mathbf{F} = \{\mathbf{f}_1,...,\mathbf{f}_N \}$ ($\mathbf{f}_i \in \{0,1\}^q$, $1 \leq  i \leq N$) where $N$ is the number of entities and $\mathbf{f}_i$ is the feature set of the entity with embedding $\mathbf{x}_i$. $p$ and $q$ denote the embeddings' dimensionality and the total number of converted binary features.

With the processed data, we describe the \textit{EmbeddingTree} algorithm with details in 
Algorithm \ref{ET}.
First, before a tree splitting, PCA is applied to the embedding data to find the first eigenvector and project all embeddings onto the principal direction. 
Next, we iterate through the $q$ features (line 6) and evaluate them based on the splitting criteria described in Section \ref{split} to pick out the best feature for splitting (line 8-10), using the feature's binary value (line 11-13).
Lastly, the above procedure is executed recursively (line 15-16) until the splitting criterion ($\Theta$), e.g., the number of entities per tree node or the tree depth, is no longer satisfied (line 2).

With the given embedding and feature data, the whole procedure is deterministic. The goal of constructing \textit{EmbeddingTree} is to provide users a tool to \textit{hierarchically explore} and understand the embeddings by incorporating the entity features. 


\begin{algorithm}[h]
  \caption{Build an \textit{EmbeddingTree}}\label{ET}
  \footnotesize
   \begin{algorithmic}[1]
   \Procedure{BuildTree}{$[\mathbf{X},\mathbf{F}], q, \Theta$}
    \If{$\Theta \hspace{1mm} is \hspace{1mm} not\hspace{1mm} satisfied$ }
    \State return LeafNode([$\mathbf{X},\mathbf{F}$])
    \Else
    \State $max\_t \leftarrow -\infty$ 
        \For{$k \in \{1,..., q\}$}
        \State $t = Embedding\textendash BIC([\mathbf{X}, \mathbf{F}^{k}])$
            \If{$t > max\_t$}
            \State $bestFea = k$ 
            \State $max\_t = t$
            \EndIf
        \EndFor
    \EndIf
    \State $[\mathbf{X},\mathbf{F}]_{left} = \{\mathbf{x} \in \mathbf{X} \vert  \mathbf{F}_{bestFea} == 0\} $\\
    \State $[\mathbf{X},\mathbf{F}]_{right} = \{\mathbf{x} \in \mathbf{X} \vert  \mathbf{F}_{bestFea} == 1\} $\\
    \State Children.Left = $BuildTree([\mathbf{X},\mathbf{F}]_{left}, q, \Theta) $
    \State Children.Right = $BuildTree([\mathbf{X},\mathbf{F}]_{right}, q, \Theta) $
    \State \Return Children
   \EndProcedure
 \end{algorithmic}
\end{algorithm}

\subsection{2-GMM Splitting with Embedding-BIC.}
\label{split}
One critical component of Algorithm~\ref{ET} is the criterion used to select the best splitting feature. The criterion is computed based on the approximate BIC for GMMs inspired by~\cite{madhyastha2020geodesic}.
Specifically, each binary feature value splits the embeddings into two clusters, each is then formulated as a Gaussian. Suppose one feature $\mathbf{F}^k$, $k \in [1, q]$ splits the first $s$ entities with feature value $0$ and the rest $N-s$ entities with feature value 1, the complete likelihood is:
\begin{equation}
\label{eq1}
P(x,\mu,\sigma,w )= \prod_{i=1}^s w_1 \mathcal{N}(x_i;\mu_1,{\sigma_1}^2)\prod_{i=s+1}^N w_2 \mathcal{N}(x_i;\mu_2,{\sigma_2}^2),
\end{equation}
where $\mathcal{N}$ is the probability density function of Gaussian distribution. For $i \in (1,2)$, $\mu_i$ is the mean,  $\sigma_i$ is the standard deviation of the Gaussian functions, $w_i$ is the weight of the cluster and $x_i$ is the PCA projected value of the embedding $\mathbf{x}_i$. 

We estimate the means ($\hat{\mu}_i$), variances ($\hat{\sigma}_i$) and weights ($\hat{w}_i$) for both clusters using the maximum likelihood estimation (MLE).
\vspace{-2mm}
\[
\hat{\mu_1} = \frac{1}{s}\sum_{i = 1}^s x_i,\hspace{5mm}
\hat{\sigma_1} = \frac{1}{s}\sum_{i = 1}^s ||x_i - \hat{\mu_1}||^2, \hspace{5mm}
\hat{w_1}= \frac{s}{N},
\]
\vspace{-2mm}
\[
\hat{\mu_2} = \frac{1}{N-s}\sum_{i=s+1}^N x_i,\hspace{5mm}
\hat{\sigma_2} = \frac{1}{N-s}\sum_{i=s+1}^N ||x_i - \hat{\mu_2}||^2, \hspace{5mm}
\hat{w_2}= \frac{N-s}{N}.
\]
\vspace{-2mm}

Plugging in the MLE for all the parameters, 
substituting the probability density function of Gaussian $\mathcal{N}=\frac{1}{\sqrt{2\pi}\sigma}\exp^{-\frac{(x-\mu)^2}{2 \sigma^2}}$ into Equation~\ref{eq1} and simplifying the outcome, we get the following expression for the log likelihood for any given $s$:
\begin{multline}
\hat{L}_s = -\frac{s}{2}\log2\pi\hat{\sigma_1}^2-\frac{N-s}{2}\log2\pi\hat{\sigma_2}^2 \\
+ s\log\hat{w_1}+(N-s)\log\hat{w_2}.
\end{multline}
We call our algorithm \textit{Embedding-BIC} algorithm. Specifically, it chooses the feature as the split-point that maximizes $\hat{L_s}$ . The detailed explanation and pseudocode for this algorithm can be found in our Appendix. 


\subsection{Embedding and Entity Feature Consistency}
\textit{EmbeddingTree} splits the whole dataset (embedding data + feature data) into different clusters represented by the leaf nodes.  The relationship between entity features and embeddings can be classified into two categories, \textit{consistent} and \textit{inconsistent}.

If entity features and embeddings are consistent, the embeddings fully encode the feature information, and they can be perfectly separated by the \textit{EmbeddingTree}, i.e., each leaf node contains a single cluster of embeddings. In the second case, either the embedding doesn't show the cluster information under a subset of features or there are multiple clusters in one leaf node. The user can dive into the problematic clusters to diagnose the reasons for the inconsistency. 

\section{Visualization Tool}
\label{sec:visualization}

We develop a visualization system to help users interactively explore the generated \textit{EmbeddingTree}. The system contains three main visualization components, as shown in Figure~\ref{fig:vis}.

The \textbf{Tree view} (Figure~\ref{fig:vis}a) presents the structure of the \textit{EmbeddingTree}. The texts of internal nodes denote the splitting condition for each branch of the tree, whereas the numbers in tree leaves represent the number of entities falling into the corresponding leaves. The links connecting a parent and a child node show the flowing of data entities (branched by different splitting conditions), and their width is proportional to the number of entities.

For the exploration of large and deep trees, we empower the tree visualization with interactive collapsing of the tree nodes. As shown in Figure~\ref{fig:vis}-a1, clicking individual internal nodes will expand/fold the corresponding tree branches to save the screen space (e.g., the light blue nodes have been folded). Clicking leaf nodes will select the entities of the corresponding tree branch into the Dimensionality Reduction view for detailed exploration.

The \textbf{Dimensionality Reduction view} (Figure~\ref{fig:vis}b) demonstrates the selected data entities by projecting their HD embedding into 2D for visualization and interaction. The selected leaf in Figure~\ref{fig:vis}a (highlighted with the red dashed circle) has 471 entities, and they are represented by the 471 points in the Dimensionality Reduction view. We have explored the three most popular dimensionality reduction algorithms, i.e., tSNE, PCA, and UMAP. All meet our requirement of disclosing the cluster pattern from the selected data entities. We choose PCA as the default algorithm considering its significantly lower computational cost. With the visually identified clusters from this view, one can flexibly select the corresponding data entities with lasso-selections (Figure~\ref{fig:vis}-b1) and investigate detailed entity features in the Data Table view.

The \textbf{Data Table view} (Figure~\ref{fig:vis}c) shows the raw data entities in a 2D table, with rows representing individual entities and columns representing individual entity features and embedding dimensions. From the table header, users can conveniently search or filter data entities. They can also sort them by a given feature of interest.

\label{sec:experiment}
\section{Case Studies}
This section describes two case studies with embeddings trained from credit-card transactions and the public 30Music data to show the efficacy of our \textit{EmbeddingTree} algorithm and the visualizations.

\begin{figure*}[tbh]
    \centering
  \includegraphics[width=\textwidth]{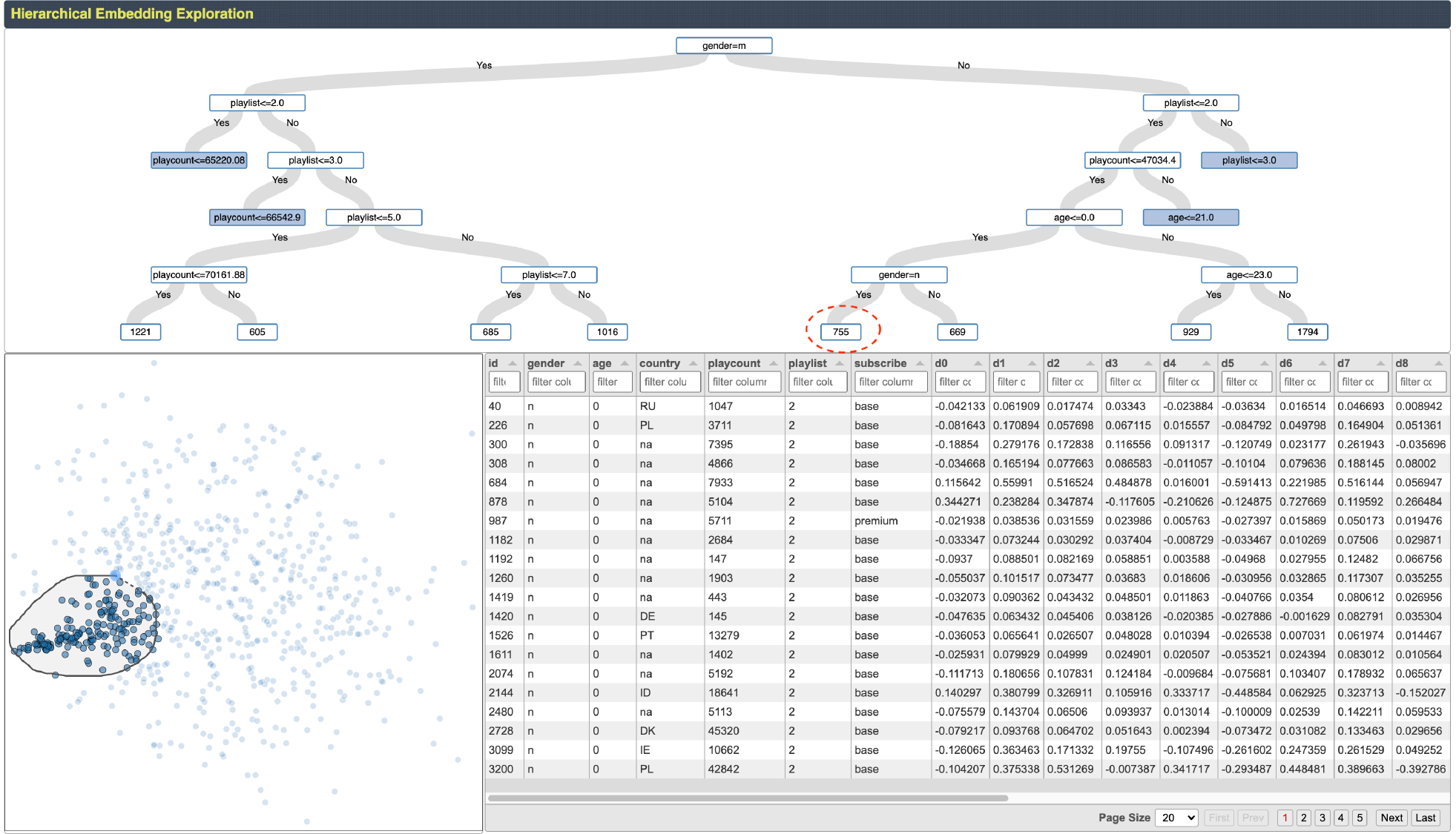}
  \caption{\textit{EmbeddingTree} for the embedding of the 30Music data. Entities of the selected node show a single cluster with random distribution.}
  \label{fig:music}
\end{figure*}

\subsection{Merchant Embedding from Transaction Data}
\label{sec:merchant}
Merchant embedding plays a significant role in many payment-related applications, such as restaurant recommendation \cite{du2019pcard} and merchant identification~\cite{yeh2020merchant}.
The embedding dataset in this study is generated using a real-world transaction dataset involving $70$ million merchants and $260$ million customers. 
The merchant embedding is generated by Word2vec~\cite{MSC13,du2019}, where each merchant is treated as a word and each customer as a document. 
For this experiment, we focus on the embeddings for merchants from four cities (city=0, 1, 2, 3): San Francisco(city=0), Los Angeles (city=1), Chicago (city=2), and Manhattan (city=3). We choose the three most frequent merchant category codes (MCC) to filter the embeddings, which are fast food (MCC=5814), gas station (MCC=5411), and grocery (MCC=5499). After filtering, the numbers of merchants in the four cities are $9926$, $21456$, $19958$ and $21202$, respectively. 
Besides \textit{location} and \textit{MCC}, the third feature we are given is the number of transactions of the merchant (\textit{frequency}). Since it is not a categorical feature, we've binned it into buckets based on the percentiles (below $33\%$, $33\%{\sim}66\%$, and above $66\%$).
The merchant \textit{EmbeddingTree} is built with Algorithm \ref{ET} using the splitting criteria introduced in Section \ref{split}. The resulting tree is shown in Figure~\ref{fig:vis}a.


From the visualization in the Tree view, we can conclude that \textit{location} is the most important feature since the tree uses it repetitively in early levels to split the embedding data. However, under the same city, the features that further distinguish the entities/merchants are different. For example, when \textit{city=1} (the first left branch in Figure~\ref{fig:vis}a), \textit{MCC} is the dominant feature, but it is \textit{frequency} when \textit{city=2} (the corresponding node was collapsed in Figure~\ref{fig:vis}a).
This observation demonstrates the strength of the hierarchical exploration compared to the parallel exploration. 
The proposed hierarchical exploration allows a localized feature importance analysis, and such analysis is not possible with parallel exploration.

We can also observe the data inconsistency issue from our visualization tool. The selected leaf node contains $471$ merchants, representing \textit{MCC=5499}, \textit{City=1}, and \textit{frequency} between $73$ and $101.5$. The embedding should be well clustered if using only the merchants' features to do the embedding. However, we can clearly see the two clusters in Figure~\ref{fig:vis}b. By analyzing the details of the two clusters, it is interesting to find that most of the merchants on the left cluster are Square merchants, which accept and process payment with Square. The merchants on the right are non-Square merchants. The visualization deepens users' understanding of the embeddings and helps to discover semantics hidden in the embeddings.

Additionally, by deploying the visualization tool to be used by domain practitioners, they notice that our tool can help embedding generation for unseen entities. Specifically, most embeddings are pre-computed and their generation relies on the rich interactions between a large number of data entities. When using the embeddings for downstream tasks, new entities (that were not seen in the pre-computation stage) may occur, causing the failure of the pre-trained embeddings. Our \textit{EmbeddingTree} can help to handle this case.
For example, a newly opened merchant only has a few transactions. Generating embeddings for such a merchant is very difficult due to the lack of historical information and interactions with others.
By traversing the \textit{EmbeddingTree} built for the pre-computed merchant embeddings, we can quickly locate the embedding cluster that the new merchant should fall into, using its feature values (e.g. merchant type and location). The mean of all embeddings in the cluster can be considered as the initial embedding of the new merchant. In short, \textit{EmbeddingTree} helps the new merchant find the group it is most similar to based on its feature information.



\subsection{User/Track Embedding for Music Recommendation}
\label{sec:music}
The embedding examined in this section is learned from the 30Music dataset~\cite{turrin201530music}.
The dataset consists of the listening behavior of $45{,}167$ users and their meta information, such as \textit{gender}, \textit{age}, \textit{country}, \textit{play count}, \textit{playlist count}, and \textit{subscribe type}.
The dataset also has $5{,}675{,}143$ tracks of music, and the total number of user-track pairs is $31{,}351{,}945$.
We leverage the listening relationship, i.e., the user-track pairs, to learn the embedding for individual users using the framework proposed in~\cite{yeh2020towards}. 

The \textit{EmbeddingTree} built for the resulting users' embedding is shown in Figure~\ref{fig:music}. From the hierarchy, the \textit{gender} of a user is the most significant feature differentiating their listening behaviors (encoded in the embedding). Following the top-down exploration, the \textit{EmbeddingTree} is further split by \textit{playlist}, followed by \textit{play count}.

Investigating users from one leaf node (highlighted in the red circle in Figure~\ref{fig:music}), we found that the users' embedding shows a single cluster in the Dimensionality Reduction view with no obvious outlier. However, as shown in the Data Table view, the \textit{country} and \textit{subscribe type} of the users have different values.
This reveals a potential deficiency of the learned embeddings, i.e., they fail to capture the variance from these two features.
To verify this, one can check if the corresponding feature values are correct or not. For example, the user with id $987$ has a \textit{premium} subscribe type, which differs from all other users that have the \textit{base} type. If the data source is available, a quick check of this user can help to perform data quality control.
If there is no data quality issue, the \textit{subscribe type} is probably not properly encoded in the embedding. The \textit{country} feature can be explored similarly. The exploration here helps to disclose potential issues of the learned embeddings, which motivates users to explicitly inject less-encoded features into the embedding learning process to refine the embedding.


   


\vspace{-2mm}
\section{Conclusion and Future Works}
\label{sec:conclusion} 
In this paper, we introduce \textit{EmbeddingTree}, an approach that allows users to hierarchically organize embeddings according to their entity features. 
The approach helps to analyze the importance of entity features and understand the relationship between features in the learned embeddings. A visualization tool is developed and tailored for the visualization and exploration of hierarchical embeddings.
We demonstrate the utility of the visualization tool with two use cases. One discovers minority features embedded in the merchant embedding data, the other discloses the feature and embedding inconsistency in the embedding of the 30Music data. Our algorithm and the visualization tool have been used by domain practitioners to help their daily practices.  
In the future, we would like to conduct statistical user studies to comprehensively validate the effectiveness of our system. Also, we plan to enable users to inject their domain knowledge of different data features into the \textit{EmbeddingTree} construction process by manually growing/pruning the tree.
Lastly, we will work with more embedding data to further optimize the generalizability and scalability of our algorithm and visualizations.

\ifpdf
  \pdfoutput=1\relax                   
  \pdfcompresslevel=9                  
  \pdfoptionpdfminorversion=7          
  \DeclareGraphicsExtensions{.pdf,.png,.jpg,.jpeg} 
\else
  \ExecuteOptions{dvips}
  \usepackage{graphicx}                
  \DeclareGraphicsExtensions{.eps}     
\fi%

\graphicspath{{figures/}{pictures/}{images/}{./}} 

\renewcommand*\ttdefault{txtt}         

\setlist[itemize]{leftmargin=*, topsep=0pt, itemsep=0pt, parsep=0pt, partopsep=0pt}
\setlist[enumerate]{leftmargin=*, topsep=0pt, itemsep=0pt, parsep=0pt, partopsep=0pt}

\onlineid{0}
\vgtccategory{Research}


\appendix
\section{Appendix}
\subsection{Details of Section 3.2: 2-GMM Splitting with Embedding-BIC.}
\label{split}
One critical component of the Algorithm in Section 3.1 is the criterion used to select the best splitting feature. The criterion is computed based on the Fast-BIC for GMMs inspired by [29].

We assume the embedding after PCA projection can be modeled as two mixture Gaussians. The expectation-maximization (EM) algorithm is used to jointly estimate all the parameters and latent variables. The latent variables, $z_{i,j}$, denote the probability that sample $i$ is in cluster $j$.
With $N$ as the number of observations and $J$ be the number of Gaussian clusters (in this case, $J = 2$), $z = \{z_{1,1}, z_{1,2}, . . . ,z_{N, J-1}, z_{N, J} \}$, the complete likelihood (including the latent variables) is:
\begin{equation}
P(x,\mu,\sigma,w,z )= \prod_{i=1}^N \prod_{j=1}^J \{w_j \mathcal{N}(x_n;\mu_j,\sigma_j^2)\}^{z_{i,j}}
\end{equation}

To find the best binary feature that splits the embedding and forms the best GMM, we go through every feature. Each candidate binary feature splits the embeddings into two clusters, each is then formulated as a Gaussian. For each feature, suppose the first $s$ embeddings have feature value $F^k=0$ and the rest $N-s$ embeddings have feature value $F^k=1$. 
We estimate the weights, means, and variances for both clusters using the maximum likelihood estimation (MLE).

\[
\hat{\mu_1} = \frac{1}{s}\sum_{i = 1}^s x_i,\hspace{5mm}
\hat{\sigma_1} = \frac{1}{s}\sum_{i = 1}^s ||x_i - \hat{\mu_1}||^2, \hspace{5mm}
\hat{w_1}= \frac{s}{N},
\]
\[
\hat{\mu_2} = \frac{1}{N-s}\sum_{i=s+1}^N x_i,\hspace{5mm}
\hat{\sigma_2} = \frac{1}{N-s}\sum_{i=s+1}^N ||x_i - \hat{\mu_2}||^2, \hspace{5mm}
\hat{w_2}= \frac{N-s}{N}.
\]

In other words, rather than the soft clustering of GMM, our algorithm performs a hard clustering. Thus, if $x_i$ is in cluster $j$, then $z_{i,j} = 1$ and $z_{i,j'} = 0$ for all $j \neq j'$. Given this approximation, the likelihood can be obtained by summing over the $z$:
\begin{equation}
P(x,\mu,\sigma,w)= \sum_z \prod_{i=1}^N \prod_{j=1}^J \{w_j \mathcal{N}(x_n;\mu_j,\sigma_j^2)\}^{z_{i,j}}
\end{equation}

Note that $z_{(i\in(0,s],j=1)}=z_{(i\in[s+1,N),j=2)}=1$ and $z_{i,j}=0$, otherwise, the above equation can be simplified to:

\begin{equation}
P(x,\mu,\sigma,w )= \prod_{i=1}^s w_1 \mathcal{N}(x_i;\mu_1,{\sigma_1}^2)\prod_{i=s+1}^N w_2 \mathcal{N}(x_i;\mu_2,{\sigma_2}^2).
\end{equation}

Plugging in the MLE for all the parameters, the maximum log-likelihood function $\hat{L}=\log P(x,\hat{\mu},\hat{\sigma},\hat{w})$ is:
\begin{multline}
    \hat{L} = \sum_{i=1}^s[\log\hat{w_1}+\log\mathcal{N}(x_i;\hat{\mu_1},\hat{\sigma_1}^2)]\\
    +\sum_{i=s+1}^N[\log\hat{w}_2+\log\mathcal{N}(x_i;\hat{\mu_2},\hat{\sigma_2}^2)].
\end{multline}
Substituting the probability density function of Gaussian $\mathcal{N}=\frac{1}{\sqrt{2\pi}\sigma}\exp^{-\frac{(x-\mu)^2}{2 \sigma^2}}$ into the above equation and simplifying, we get the following expression for the log-likelihood for any given $s$:
\begin{multline}
\hat{L}_s = -\frac{s}{2}\log2\pi\hat{\sigma_1}^2-\frac{N-s}{2}\log2\pi\hat{\sigma_2}^2 \\
+ s\log\hat{w_1}+(N-s)\log\hat{w_2},
\end{multline}
in which we have dropped terms that are not functions of the parameters.
We call our algorithm Embedding-BIC algorithm. Specifically, it chooses the feature as the split-point that maximizes $\hat{L_s}$. The detailed pseudocode for this algorithm can be found in Algorithm 1. 

\label{APPBIC}
\begin{algorithm}[h]
\caption{Calculate the BIC score for embedding $\mathbf{X}$ with a given binary feature $\mathbf{F}^k$.}
\label{BIC}
\begin{algorithmic}[1]
\Procedure{Embedding-BIC}{$\mathbf{X}, \mathbf{F}^k)$}
\For{$i \in [1,N]$}
  \State $x_i = PCA(\mathbf{x}_i)$
  \State $C_1 \leftarrow \mathbf{f}_i^k = 0$
  \State $C_2 \leftarrow \mathbf{f}_i^k = 1$
  \For{$j \in \{1,2\}$} 
    \State $n_j= length(C_j)$
    \State $\hat{w_j} =n_j/n$
    \State $\hat{\mu}_j = \frac{1}{n_j}\sum_{x_i \in C_j} x_i$ 
    \State $\hat{\sigma_j}^2 = \frac{1}{n_j}\sum_{x_i \in C_j}(x_i - \hat{\mu_j})^2$ 
   \EndFor
   \State BIC\_curr $= 
    -\frac{n_1}{2}\log 2\pi \hat{\sigma}_1^2 
    -\frac{n_2}{2}\log 2\pi \hat{\sigma}_2^2 
    + n_1 \log \hat{w}_1 
    + n_2 \log \hat{w}_1 $
  \EndFor
  \State \Return BIC\_curr 
  \EndProcedure
 \end{algorithmic}
\end{algorithm}

\bibliographystyle{IEEEtran}
\bibliography{paper}

\begin{thebibliography}{10}
\providecommand{\url}[1]{#1}
\csname url@samestyle\endcsname
\providecommand{\newblock}{\relax}
\providecommand{\bibinfo}[2]{#2}
\providecommand{\BIBentrySTDinterwordspacing}{\spaceskip=0pt\relax}
\providecommand{\BIBentryALTinterwordstretchfactor}{4}
\providecommand{\BIBentryALTinterwordspacing}{\spaceskip=\fontdimen2\font plus
\BIBentryALTinterwordstretchfactor\fontdimen3\font minus
  \fontdimen4\font\relax}
\providecommand{\BIBforeignlanguage}[2]{{%
\expandafter\ifx\csname l@#1\endcsname\relax
\typeout{** WARNING: IEEEtran.bst: No hyphenation pattern has been}%
\typeout{** loaded for the language `#1'. Using the pattern for}%
\typeout{** the default language instead.}%
\else
\language=\csname l@#1\endcsname
\fi
#2}}
\providecommand{\BIBdecl}{\relax}
\BIBdecl

\bibitem{mikolov2013efficient}
T.~Mikolov, K.~Chen, G.~Corrado, and J.~Dean, ``Efficient estimation of word
  representations in vector space,'' \emph{arXiv:1301.3781}, 2013.

\bibitem{camacho2018word}
J.~Camacho-Collados and M.~T. Pilehvar, ``From word to sense embeddings: A
  survey on vector representations of meaning,'' \emph{JAIR}, vol.~63, pp.
  743--788, 2018.

\bibitem{tshitoyan2019unsupervised}
V.~Tshitoyan, J.~Dagdelen, L.~Weston, A.~Dunn, Z.~Rong, O.~Kononova, K.~A.
  Persson, G.~Ceder, and A.~Jain, ``Unsupervised word embeddings capture latent
  knowledge from materials science literature,'' \emph{Nature}, vol. 571, no.
  7763, p.~95, 2019.

\bibitem{grbovic2018real}
M.~Grbovic and H.~Cheng, ``Real-time personalization using embeddings for
  search ranking at airbnb,'' in \emph{KDD}, 2018, pp. 311--320.

\bibitem{wang2018billion}
J.~Wang, P.~Huang, H.~Zhao, Z.~Zhang, B.~Zhao, and D.~L. Lee, ``Billion-scale
  commodity embedding for e-commerce recommendation in alibaba,'' in
  \emph{KDD}, 2018, pp. 839--848.

\bibitem{zhao2018learning}
X.~Zhao, R.~Louca, D.~Hu, and L.~Hong, ``Learning item-interaction embeddings
  for user recommendations,'' \emph{arXiv:1812.04407}, 2018.

\bibitem{du2019}
M.~Du, R.~Christensen, W.~Zhang, and F.~Li, ``Pcard: Personalized restaurants
  recommendation from card payment transaction records,'' in \emph{WWW}, 2019,
  pp. 2687--2693.

\bibitem{smilkov2016embedding}
D.~Smilkov, N.~Thorat, C.~Nicholson, E.~Reif, F.~B. Vi{\'e}gas, and
  M.~Wattenberg, ``Embedding projector: Interactive visualization and
  interpretation of embeddings,'' \emph{arXiv preprint arXiv:1611.05469}, 2016.

\bibitem{heimerl2020embcomp}
F.~Heimerl, C.~Kralj, T.~Moller, and M.~Gleicher, ``embcomp: Visual interactive
  comparison of vector embeddings,'' \emph{IEEE Transactions on Visualization
  and Computer Graphics}, 2020.

\bibitem{liu2019latent}
Y.~Liu, E.~Jun, Q.~Li, and J.~Heer, ``Latent space cartography: Visual analysis
  of vector space embeddings,'' in \emph{Computer Graphics Forum}, vol.~38,
  no.~3.\hskip 1em plus 0.5em minus 0.4em\relax Wiley Online Library, 2019, pp.
  67--78.

\bibitem{liu2017visual}
S.~Liu, P.-T. Bremer, J.~J. Thiagarajan, V.~Srikumar, B.~Wang, Y.~Livnat, and
  V.~Pascucci, ``Visual exploration of semantic relationships in neural word
  embeddings,'' \emph{IEEE transactions on visualization and computer
  graphics}, vol.~24, no.~1, pp. 553--562, 2017.

\bibitem{berger2020visually}
M.~Berger, ``Visually analyzing contextualized embeddings,'' in \emph{2020 IEEE
  Visualization Conference (VIS)}.\hskip 1em plus 0.5em minus 0.4em\relax IEEE,
  2020, pp. 276--280.

\bibitem{wang2021constrained}
Y.~Wang, Y.~Zheng, Y.~Peng, M.~Yeh, Z.~Zhuang, D.~Mahashweta, B.~Mangesh,
  F.~Li, W.~Zhang, and J.~M. Phillips, ``Constrained non-affine alignment of
  embeddings,'' in \emph{2021 IEEE International Conference on Data Mining
  (ICDM)}.\hskip 1em plus 0.5em minus 0.4em\relax IEEE, 2021, pp. 1403--1408.

\bibitem{MSC13}
T.~Mikolov, I.~Sutskever, K.~Chen, G.~S. Corrado, and J.~Dean, ``Distributed
  representations of words and phrases and their compositionality,'' vol.~26,
  2013.

\bibitem{pennington2014glove}
J.~Pennington, R.~Socher, and C.~D. Manning, ``Glove: Global vectors for word
  representation,'' in \emph{Empirical Methods in Natural Language Processing
  (EMNLP)}, 2014, pp. 1532--1543.

\bibitem{harris1954}
Z.~Harris, ``Distributional structure,'' \emph{Word}, vol.~10, no.~23, pp.
  146--162, 1954.

\bibitem{li2019learning}
B.~Li and D.~Pi, ``Learning deep neural networks for node classification,''
  \emph{Expert Systems with Applications}, vol. 137, pp. 324--334, 2019.

\bibitem{fan2019graph}
W.~Fan, Y.~Ma, Q.~Li, Y.~He, E.~Zhao, J.~Tang, and D.~Yin, ``Graph neural
  networks for social recommendation,'' in \emph{WWW}, 2019, pp. 417--426.

\bibitem{wei2017cross}
X.~Wei, L.~Xu, B.~Cao, and P.~S. Yu, ``Cross view link prediction by learning
  noise-resilient representation consensus,'' in \emph{WWW}, 2017, pp.
  1611--1619.

\bibitem{cai2018comprehensive}
H.~Cai, V.~W. Zheng, and K.~C.-C. Chang, ``A comprehensive survey of graph
  embedding: Problems, techniques, and applications,'' \emph{IEEE Transactions
  on Knowledge and Data Engineering}, vol.~30, no.~9, pp. 1616--1637, 2018.

\bibitem{goyal2018graph}
P.~Goyal and E.~Ferrara, ``Graph embedding techniques, applications, and
  performance: A survey,'' \emph{Knowledge-Based Systems}, vol. 151, pp.
  78--94, 2018.

\bibitem{rahman2019fairwalk}
T.~Rahman, B.~Surma, M.~Backes, and Y.~Zhang, ``Fairwalk: Towards fair graph
  embedding,'' in \emph{Proceedings of the 2019 International Joint Conferences
  on Artifical Intelligence (IJCAI)}, 2019.

\bibitem{wang2017knowledge}
Q.~Wang, Z.~Mao, B.~Wang, and L.~Guo, ``Knowledge graph embedding: A survey of
  approaches and applications,'' \emph{IEEE Transactions on Knowledge and Data
  Engineering}, vol.~29, no.~12, pp. 2724--2743, 2017.

\bibitem{hamilton2017inductive}
W.~Hamilton, Z.~Ying, and J.~Leskovec, ``Inductive representation learning on
  large graphs,'' \emph{Advances in neural information processing systems},
  vol.~30, 2017.

\bibitem{mikolov2013distributed}
T.~Mikolov, I.~Sutskever, K.~Chen, G.~S. Corrado, and J.~Dean, ``Distributed
  representations of words and phrases and their compositionality,'' in
  \emph{Advances in neural information processing systems}, 2013, pp.
  3111--3119.

\bibitem{yeh2020towards}
C.-C.~M. Yeh, D.~Gelda, Z.~Zhuang, Y.~Zheng, L.~Gou, and W.~Zhang, ``Towards a
  flexible embedding learning framework,'' in \emph{2020 International
  Conference on Data Mining Workshops (ICDMW)}.\hskip 1em plus 0.5em minus
  0.4em\relax IEEE, 2020, pp. 605--612.

\bibitem{liu2020latentvis}
X.~Liu and J.~Wang, ``Latentvis: Investigating and comparing variational
  auto-encoders via their latent space.'' in \emph{CIKM Workshops}, 2020.

\bibitem{li2018embeddingvis}
Q.~Li, K.~S. Njotoprawiro, H.~Haleem, Q.~Chen, C.~Yi, and X.~Ma,
  ``Embeddingvis: A visual analytics approach to comparative network embedding
  inspection,'' in \emph{2018 IEEE Conference on Visual Analytics Science and
  Technology (VAST)}.\hskip 1em plus 0.5em minus 0.4em\relax IEEE, 2018, pp.
  48--59.

\bibitem{boggust2022embedding}
A.~Boggust, B.~Carter, and A.~Satyanarayan, ``Embedding comparator: Visualizing
  differences in global structure and local neighborhoods via small
  multiples,'' in \emph{27th international conference on intelligent user
  interfaces}, 2022, pp. 746--766.

\bibitem{madhyastha2020geodesic}
M.~Madhyastha, G.~Li, V.~Strnadov{\'a}-Neeley, J.~Browne, J.~T. Vogelstein,
  R.~Burns, and C.~E. Priebe, ``Geodesic forests,'' in \emph{KDD}, 2020, pp.
  513--523.

\bibitem{du2019pcard}
M.~Du, R.~Christensen, W.~Zhang, and F.~Li, ``Pcard: Personalized restaurants
  recommendation from card payment transaction records,'' in \emph{WWW}, 2019,
  pp. 2687--2693.

\bibitem{yeh2020merchant}
C.-C.~M. Yeh, Z.~Zhuang, Y.~Zheng, L.~Wang, J.~Wang, and W.~Zhang, ``Merchant
  category identification using credit card transactions,'' in \emph{IEEE
  International Conference on Big Data}, 2020, pp. 1736--1744.

\bibitem{turrin201530music}
R.~Turrin, M.~Quadrana, A.~Condorelli, R.~Pagano, and P.~Cremonesi, ``30music
  listening and playlists dataset.'' in \emph{RecSys Posters}, 2015.

\end{thebibliography}
\end{document}